\documentclass[conference]{IEEEtran}
\usepackage[square,numbers]{natbib} % has a nice set of citation styles and commands
\let\cite\citep
\IEEEoverridecommandlockouts

\usepackage{mathtools} % amsmath with fixes and additions
\usepackage{booktabs} % commands to create good-looking tables
\usepackage{tikz} % nice language for creating drawings and diagrams
\usepackage{array}
\usepackage{amsmath}
\usepackage{amsthm}
\usepackage{amssymb}
\usepackage{graphicx}
\usepackage{subfigure}
\usepackage{xspace}
\usepackage{paralist}
\usepackage{multirow}
\usepackage{caption,latexsym}
\usepackage[ruled,vlined,oldcommands]{algorithm2e}
\usepackage{wrapfig}
\usepackage{url}

\newtheorem{definition}{Definition}

\newtheorem{lemma}{Lemma}

\usepackage{fancyhdr}
\newcommand{\eat}[1]{}

\begin{document}
% \fancyhf{}
% \renewcommand{\headrulewidth}{0pt}
% \fancyfoot[c]{}
% \fancypagestyle{FirstPage}{
% \lfoot{978-1-6654-8045-1/22/\$31.00 \copyright2022 IEEE} 
% }

\title{Verifying Relational Explanations: A Probabilistic Approach \thanks{This research was supported by NSF awards \#2008812 and \#1934745. The opinions, findings, and results are solely the authors' and do not reflect those of the funding agencies.}
}

\IEEEoverridecommandlockouts
\IEEEpubid{\makebox[\columnwidth]{979-8-3503-2445-7/23/\$31.00 ©2023 IEEE \hfill} 
 \hspace{\columnsep}\makebox[\columnwidth]{ }}

% https://colab.research.google.com/drive/1eqU3KX9krwM0toy-SW4KrPSdrDHdWFxy?usp=sharing

 \author{
     \IEEEauthorblockN{Abisha Thapa Magar\IEEEauthorrefmark{1}, Anup Shakya\IEEEauthorrefmark{1}, Somdeb Sarkhel\IEEEauthorrefmark{2}, Deepak Venugopal\IEEEauthorrefmark{1}}
     \IEEEauthorblockA{\IEEEauthorrefmark{1}University of Memphis
     \\\{thpmagar, ashakya, dvngopal\}@memphis.edu}
     \IEEEauthorblockA{\IEEEauthorrefmark{2}Adobe Research   
     \\\{sarkhel\}@adobe.com}
 }

\maketitle
\IEEEpubidadjcol
\begin{abstract}
    Explanations on relational data are hard to verify since the explanation structures are more complex (e.g. graphs). To verify interpretable explanations (e.g. explanations of predictions made in images, text, etc.), typically human subjects are used since it does not necessarily require a lot of expertise. However, to verify the quality of a relational explanation requires expertise and is hard to scale-up. GNNExplainer is arguably one of the most popular explanation methods for Graph Neural Networks. In this paper, we develop an approach where we assess the uncertainty in explanations generated by GNNExplainer. Specifically, we ask the explainer to generate explanations for several counterfactual examples. We generate these examples as symmetric approximations of the relational structure in the original data. From these explanations, we learn a factor graph model to quantify uncertainty in an explanation. Our results on several datasets show that our approach can help verify explanations from GNNExplainer by reliably estimating the uncertainty of a relation specified in the explanation.
\end{abstract}

\section{Introduction}

Relational data is ubiquitous in nature. Healthcare records, social networks, biological data and educational data are all inherently relational in nature. Graphs are the most common representations for relational data and Graph Neural Networks (GNNs)~\cite{GCN_Kipf,gat_2018} are arguably among the most popular approaches used for learning from graphs.

%Popular Graph Neural Networks (GNNs) include Graph Convolution Networks (GCNs)~\cite{GCN_Kipf} and Graph Attention Networks (GATs)~\cite{gat_2018}. At the same time, to improve trust in predictions made by the GNN, we need to be able to explain these predictions. 
%Shakya et al.~\cite{shakya23verification} present a framework for verifying embeddings in GNNs to strengthen the trustworthiness of GNN predictions.
While there have been several methods related to explainable AI (XAI)~\cite{gunning19}, it should be noted that explaining GNNs is perhaps more challenging than explaining non-relational machine learning algorithms that work on i.i.d (independent and identically distributed) data. For instance, techniques such as LIME~\cite{Lime_kdd16} or SHAP~\cite{Shap_NIPS17} explain a prediction based on interpretable features such as pixel-patches (for images) or words (for language). The quality of explanations produced using such methods are generally verified with human subjects since typically, anyone can understand the explanations that are produced and thus can judge their quality. However, in the case of GNNs, explanations are much more complex and cannot be verified easily through human subjects. Specifically, consider GNNExplainer~\cite{NEURIPS2019_d80b7040} arguably one of the most widely used explainers for GNNs. Given a relational graph where the task is to classify nodes in the graph, GNNExplainer produces a subgraph as the explanation for a node prediction. Clearly, it is very hard to verify such an explanation using human subjects since the explanation is quite abstract. If the ground truth for the explanation is known in the form of graph structures, then it is easy to verify a relational explanation. However, this does not scale up since considerable domain expertise may be needed in this case to generate correct explanations. Therefore, in this paper, we develop a probabilistic method where we verify explanations based on how the explainer explains counterfactual examples.

The main idea in our approach is to learn a distribution over explanations for variants of the input graph and quantify uncertainty in an explanation based on this distribution. In particular, each variant can be considered as a counterfactual to the true explanation and we represent the distribution over these explanations in the form of a probabilistic graphical model (PGM). In particular, we impose a constraint where the distribution is over explanations for symmetrical counterfactual examples. Intuitively, if the input to the explainer changes, since real-world data has symmetries~\cite{sym10010029}, our distribution will be represented over more likely counterfactual examples. 

To learn such a distribution over symmetric counterfactual explanations, we perform a Boolean factorization of the relations specified in the original graph and learn low-rank approximations for them. Specifically, a low-rank approximation represents all the relationships in the data by a smaller number of Boolean patterns. To do this, it introduces symmetries into the approximated relational graph~\cite{NIPS2013_7940ab47}. We explain each of the symmetric approximations using GNNExplainer and represent the distribution over these in the form of a factor graph~\cite{fg_2004}. We calibrate this using Belief Propagation~\cite{NIPS2000_61b1fb3f} to compute the distributions over relations specified in an explanation. To quantify uncertainty in an explanation generated by GNNExplainer on the original graph, we measure the reduction in uncertainty in the calibrated factor graph when we inject knowledge of the explanation into the factor graph.

We perform experiments on several benchmark relational datasets for node classification using GCNs. In each case, we estimate the uncertainty of relations specified in explanations given by GNNExplainer. We use the McNemar's statistical test to evaluate the significance of these estimations on the model learned by the GCN. We compare our approach with the estimates of uncertainty that are directly provided by GNNExplainer.
We show that the McNemar's test reveals that using our approach to estimate the uncertainty of an explanation is statistically more reliable than using the estimates produced by GNNExplainer.

%which allows us to learn if the explanations are  
\section{Related Work}

%On the Complexity and Approximation of Binary Evidence in Lifted Inference
%Van den Broeck and Darwiche[]

%Explaining Explanations in AI
%Mittelstadt et at[]

% Generative Causal Explanations for Graph Neural Networks (ICML'21)

% Higher-Order Explanations of Graph Neural Networks via Relevant Walks (IEEE Transactions on Pattern Analysis and Machine Intelligence '22)

% CF-GNNExplainer: Counterfactual Explanations for Graph Neural Networks (AISTATS'22)

% Probing GNN Explainers: A Rigorous Theoretical and Empirical Analysis of GNN Explanation Methods (AISTATS'22)
% When Comparing to Ground Truth is Wrong: On Evaluating GNN Explanation Methods (KDD'21)

% \subsection{Explanation Methods}
Mittelstadt et al.~\cite{10.1145/3287560.3287574} compare the emerging field of explainable AI (XAI)  with what explanations mean in other fields such as social sciences, philosophy, cognitive science or law. Typically it has been shown that humans psychologically prefer counterfactual explanations~\cite{MILLER20191}. 
% In XAI, explanations are typically of the form where they add transparency to``black-box'' models since it is hard to explain the behavior of such models. 
% In GNNs, explanations can range from attribution \cite{NEURIPS2020_417fbbf2}, feature importance, and causal explanation \cite{pmlr-v139-lin21d} to counterfactual explanations \cite{pmlr-v151-lucic22a}.
%Luo et al. \cite{pg_explainer_nips20} introduce PGExplainer which parameterizes the process of generating explanations to improve the generalizability of explanations.
Schnake et al. \cite{9547794} show a novel way to naturally explain GNNs by identifying groups of edges contributing to a prediction using higher-order Taylor expansion. GraphLIME~\cite{9811416}, an extension of the LIME framework designed for graph data, is another popular explanation method that attributes the prediction result to specific nodes and edges in the local neighborhood.  Shakya et al.~\cite{shakya23verification} present a framework that verifies semantic information in GNN embeddings. Vu et al. \cite{pgm_explainer_nips20} present PGM-Explainer which can generate explanations in the form of a PGM, where the dependencies in the explained features are demonstrated in terms of conditional probabilities. There is also a lot of research work on evaluating the explanations of these explainers. Faber et al. \cite{10.1145/3447548.3467283}  argue that the current explanation methods cannot detect ground truth and they propose three novel benchmarks for evaluating explanations. Sanchez-Lengeling et al.~\cite{NEURIPS2020_417fbbf2} present a systematic way of evaluating explanation methods with properties like accuracy, consistency, faithfulness, and stability. 
%More recently, Shakya et al.~\cite{shakya23verification} present a framework for verification of GNNs through probabilistic models.
% Agrawal et al. \cite{pmlr-v151-agarwal22b} produce an extensive theoretical and empirical analysis of the state-of-the-art GNN explanation methods in terms of faithfulness, stability, and fairness preservation. They also demonstrate the upper bounds on the violation of these properties. 
%Although there has been a wide range of research in this field, the uncertainty in the explanation of the GNNs has not been studied. In this paper, we verify the relational explanations in GNN and quantify these uncertainties.

% There are plenty of research in evaluating
\section{Background}

\subsection{Graph Convolutional Networks}

Given a graph $\mathcal{G}=({\bf V}, {\bf E}, {\bf X})$ with nodes ${\bf V}$ $=$ $\{x_1\ldots x_n\}$, edges ${\bf E}$ $=$ $\{e_1\ldots e_k\}$ s.t. $e_k \in (x_i, x_j)$ and $x_i, x_j \in \bf V$ and features ${\bf X}$ $=$ $\{{\bf X}_i\}_{i=1}^n$ s.t. ${\bf X}_i \in \mathbb{R}^d$. GCN learns representations of nodes from their neighbors by using convolutional layers which is used to classify nodes. The layer-wise propagation rule is as follows:
\begin{align}
H^{(l+1)} = \sigma(\Tilde{D}^{-\frac{1}{2}} \Tilde{A}\Tilde{D}^{-\frac{1}{2}} H^{(l)} W^{(l)})
\end{align}
where $H^{(l)} \in \mathbb{R}^{n \times d}$ is the feature matrix for layer $l$, $\Tilde{A}$ is the adjacency matrix of graph $\mathcal{G}$, $\Tilde{D}_{ii} = \sum_j \Tilde{A}_{ij}$ is the degree matrix, $W^{(l)}$ is a layer-specific trainable weight matrix and $\sigma(\cdot)$ is the activation function. 

\subsection{GNNExplainer}

Given a GCN (or any GNN) $\Phi$ trained for node classification makes prediction for a single target node $Y$. GNNExplainer generates a subgraph of the computation graph as an explanation for the prediction. The objective is formulated as a minimization of the conditional entropy for the predicted node conditioned on a subgraph of the computation graph. Specifically, 
\begin{equation}
\label{eq:condentropy}
    H(Y|\mathcal{G}=\mathcal{G}_s,{\bf X}={\bf X}_s) = -\mathbb{E}_{Y|\mathcal{G}_s,{\bf X}_s}\log[P_{\Phi}(y|\mathcal{G}_s,{\bf X}_s)]
\end{equation}
where $\mathcal{G}_s$ is a subgraph of the computation graph and ${\bf X}_s$ is a subset of features.
The subgraph is obtained by retaining/removing edges/nodes from the graph. 

% If removing an edge $(x_i,x_j)$ strongly decreases probability of prediction of the target node, then $(x_i,x_j)$ is a strong counterfactual explanation for the prediction made by $\mathcal{D}$. The optimization of the conditional entropy with a continuous relaxation which allows fractional adjacency matrices can be performed by masking the original computation graph with a learnable mask. The sigmoid of the mask is then multiplied with the computation graph of the fractional adjacency matrix and a thresholding is performed to generate the final explanation subgraph.
\section{Verification of Explanations}

We develop a likelihood-based approach on top of GNNExplainer, arguably one of the most well-known approaches for explaining relational learning in DNNs, to estimate the uncertainty in an explanation. To do this, we learn a probabilistic graphical model (PGM) that encodes  relational structure of explanations. We then perform probabilistic inference over the PGM to estimate the likelihood of a specific explanation. 

%Specifically, we learn a probabilistic graphical model (PGM) to encode relationships in the explanation and use this to answer queries regarding a new explanation.

\subsection{Counterfactual Relational Explanations}

\begin{definition}
A discrete PGM is a pair ({\bf X},{\bf F}), where {\bf X} is a set of discrete random variables and ${\bf F}$ is a set of functions, $\phi\in{\bf F}$ is defined over a subset of variables referred to as being in its scope. The joint probability distribution is the normalized product of all factors. 
$$P(\bar{\bf x})=\frac{1}{Z}\prod_{\phi}\phi(\bf x)$$
where $Z$ is the normalization constant and $\phi(\bf x)$ is the value of the function when ${\bf x}$ is projected on its scope.
\end{definition}

The {\em primal graph} $G$ of a PGM is the structure of the PGM where nodes represent the discrete random variables and cliques in the graph represent the factors. An undirected PGM is also called as a {\em Markov Network}. A directed PGM is a {\em Bayesian Network} where edges represent causal links and factors represent conditional distributions, specifically, the conditional distribution of a node in $G$ given all its parents. For the purposes of generalizing notation, we can consider these as factors. However, in a Markov Network, the product of factors is not normalized and to represent a distribution, we need to normalize this with the partition function, while in a Bayesian Network the product of factors is already normalized. A {\em factor graph} is a discrete PGM represented as a bi-partite graph, where there are two types of nodes, namely, variables and factors. The edges connect variable nodes to factor nodes. The factor represents a function over the variables connected to it (the scope of the factor). Typically, the variables are connected through a logical relationship in the factor function. Each factor function has an associated {\em weight} that encodes confidence in the relationship over variables within its scope. Higher confidence in the relationship implies higher weights and vice-versa. A factor graph can be converted to an equivalent Markov network.

A {\em relational graph} $\mathcal{G}$ is a graph where nodes represent real-world entities and edges represent binary relationships between the entities. For our purposes, we assume that the relationships in $\mathcal{G}$ are not directed. Let $\Phi$ be a DNN trained for the node classification task. That is, let ${\bf V}$ $=$ $x_1\ldots x_n$ be the nodes where ${\bf X}$ $=$ $\{{\bf X}_i\}_{i=1}^n$ are their features and $\Phi$ learns to classify nodes into one of $C$ classes, $f:{\bf V}\rightarrow{C}$.

Let $E(\mathcal{G},\Phi,Y)$ denote the GNNExplainer's explanation for $\Phi$ classifying node $Y$ in $\mathcal{G}$. $E(\mathcal{G},\Phi,Y)$ is a subgraph, i.e., a set of relations/edges (we use relations and edges interchangeably since we assume binary relationships) in $\mathcal{G}$ that explains the label assigned to $Y$ by $\Phi$. We estimate the uncertainty in $E(\mathcal{G},\Phi,Y)$ based on a PGM distribution over {\em counterfactual relational explanations}. 

\begin{definition}
    Given an explanation $E(\mathcal{G},\Phi,Y)$, a counterfactual relational explanation (CRE) is $E(\hat{\mathcal{G}},\Phi,Y)$, where $\mathcal{G}$ and $\hat{\mathcal{G}}$ differ in at least one relation.
\end{definition}

\subsection{Boolean Factorization}

\begin{figure*}
    \centering
    \includegraphics[scale=0.65]{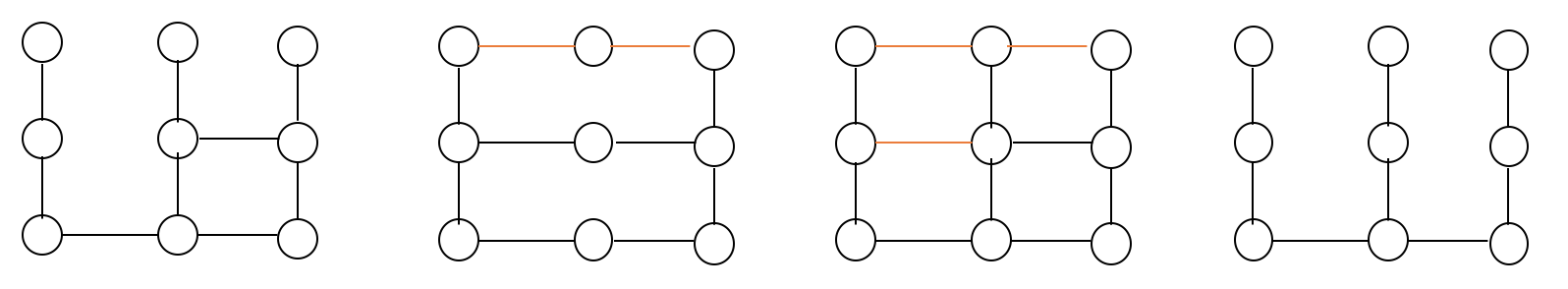}
    \caption{Given the original graph (the first one), different symmetric approximations of the original graph are shown.}
    \label{fig:exsym}
\end{figure*}

Note that computing the full set of CREs is not scalable since the size of the CRE set is exponential in the size of the relational graph. Therefore, we focus on a subset of CREs that best quantify uncertainty in the explanation. Before formalizing our approach, we illustrate this with a simple example. Consider the example shown in Fig.~\ref{fig:exsym}. To generate CREs, instead of modifying relations randomly, we add/remove relations that result in {\em symmetrical structures} as shown in the example. Thus, under the hypothesis that symmetries are ubiquitous in the real-world~\cite{sym10010029}, symmetrical CREs are likely to explain more probable counterfactual examples. Thus, a PGM over symmetrical CREs will better encode uncertainty in explanations.

Formally, let ${\bf P}$ represent all the relations in $\mathcal{G}$. We want to approximate ${\bf P}$ which can be represented as a $n\times m$ matrix using at most $k$ Boolean {\em patterns}. Specifically, the objective is as follows.
\begin{equation}
\label{eq:lrobj1}
    \arg\min_{{\bf Q}{\bf R}}|{\bf P}\ominus({\bf Q}\otimes{\bf R})|
\end{equation}
where ${\bf Q}$ is a Boolean matrix of size $n\times k$ and ${\bf R}$ is Boolean matrix of size $k\times m$ rows. The Boolean operations are defined as follows. $A\oplus B$ $=$ $A\vee B$, $A\ominus B$ $=$ $(A\wedge \neg B)\vee (\neg A\wedge B)$ and $X^{n\times m}$ $=$ $A^{n\times k}B^{k\times m}$, where $X_{ij}$ $=$ $\vee_{l=1}^kA_{il}B_{lj}$. The $l$-th column of ${\bf Q}$ and the $l$-th row of ${\bf R}$ is called as the $l$-th Boolean pattern. To solve the above optimization problem, we use Boolean Matrix Factorization (BMF). The smallest number of patterns for which we can exactly recover ${\bf P}$, i.e., the objective value is equal to 0, is known as the Boolean rank of ${\bf P}$.
It is known that computing the Boolean rank is a NP-hard problem. 
\begin{definition}
 A low-rank approximation for ${\bf P}$ with Boolean rank $r$ is a factorization with $k$ patterns such that $k<r$.
\end{definition}
Since in a low-rank approximation, we use fewer patterns than the rank, it results in a symmetric approximation of the original matrix~\cite{NIPS2013_7940ab47}. While there are several approaches for Boolean low-rank approximation~\cite{Wan_Chang_Zhao_Li_Cao_Zhang_2020}, we use a widely used approach implemented in NIMFA~\cite{JMLR:v13:zitnik12a}. Specifically, the problem is formulated as a nonlinear programming problem and solved with a penalty function algorithm. The factorization reduces the original matrix into a binary basis and mixture coefficients. By thresholding the product of the binary basis and mixture coefficient matrices, we obtain the low-rank Boolean approximation of the original matrix. Since it is hard to compute the exact rank, we use an iterative approach to obtain the set of symmetrical CREs. Specifically, for the base explanation, we perform low rank approximation with a starting rank and progressively increase the rank until the objective function in Eq.~\eqref{eq:lrobj1} is below a stopping criteria.

\subsection{Factor Graph Model}
We represent a distribution over the set of symmetrical CREs $\mathcal{S}$ using a factor graph. Specifically, each factor function represents a logical relationship between variables in the CRE. One such commonly used relationship in logic is a set of {\em Horn clauses} of the form $x_i\wedge x_j\Rightarrow Y$, where $x_i$, $x_j$ represent variables in the explanation and $Y$ is the target of the explanation. However, note that for the factor functions, we have used the logical and ($\wedge$) rather than implication ($\Rightarrow$) since the implication tends to produce uniform distributions. Specifically, whenever the head (left-side) of the horn clause is false, the clause becomes true which is not the ideal logical form for us since we want to quantify the influence of the head on the body (right hand side of the clause which is the target of explanation) when the head is true. Thus, a logical-and works better in practice. For ease of exposition, we assume that the target of an explanation has a binary class. For multi-class targets, we just create $p$ different clauses of the same form each of which corresponds to one of $p$ classes. 

Let $\{R_i\}_{i=1}^K$ be the union of all binary relations specified in the explanations in $\mathcal{S}$. The probability distribution over $\mathcal{S}$ is defined as a {\em log-linear} model as follows.
\begin{equation}
\label{eq:pgmeq}
    P(\mathcal{S}) = \frac{1}{Z}\exp(\sum_{i=1}^Kw_in(R_i,\mathcal{S}))
\end{equation}
where $n(R_i,\mathcal{S})$ is the number of true clauses of the form $x_{i1}\wedge x_{i2}\wedge T$, where $x_{i1},x_{i2}$ are entities related by $R_i$ in $\mathcal{S}$, $w_i$ is a real-valued weight for $x_{i1}\wedge x_{i2}\wedge Y$, $Z$ is the normalization constant, i.e., $\sum_{\mathcal{S}'}$ $\exp(\sum_{i=1}^Kw_in(R_i,\mathcal{S}'))$. 

To learn the weights in Eq.~\eqref{eq:pgmeq}, we maximize the likelihood over ${\bf S}\in\mathcal{S}$. Specifically,
\begin{equation}
\log\ell(\mathcal{S})=w_i\sum_{i=1}^Kn(R_i,\mathcal{S})- log Z
\end{equation}
\eat{
However,clearly $Z$ is intractable to compute. Specifically, we need to sum the probabilities over all possible explanations that can be derived from $\{R_i\}_{i=1}^K$ which is exponentially large. Therefore, using gradient descent over the weights, we obtain the following equation for the gradient,
\begin{equation}
    \frac{\partial{\ell}}{\partial{w_i}} = n(R_i,\mathcal{S}) - \mathbb{E}[n(R_i,\mathcal{S})]
\end{equation}
%We learn a factor graph from the set of symmetrical CREs as follows.
}
%It turns out that we can approximate the expectation from the maximum probability (called the MAP) solution to estimate the gradient efficiently (similar to the voted perceptron in ~\cite{singla&domingos05}). 
Since $Z$ is intractable to compute, we can use gradient descent, where $\frac{\partial{\ell}}{\partial{w_i}}$ is $n(R_i,\mathcal{S})$ $-$ $\mathbb{E}[n(R_i,\mathcal{S})]$. Similar to the voted perceptron in ~\cite{singla&domingos05}, using the current weights, we compute the max a posteriori (MAP) solution which gives us an assignment to each entity in $\mathcal{S}$. From this, we estimate $\mathbb{E}[n(R_i,\mathcal{S})]$ as follows. For each explanation ${\bf S}\in\mathcal{S}$, we check whether the clause that connects the entities corresponding to $R_i$ in ${\bf S}$ is satisfied based on the assignments in the MAP solution. $\mathbb{E}[n(R_i,\mathcal{S})]$ is the total number of satisfied clauses. Finally, we update the weights, $w_i^{(t)}$ $=$ $w_i^{(t-1)}$ $-$ $\epsilon\frac{\partial{\ell}}{\partial{w_i}}$, where $\epsilon$ is the learning rate. However, it turns out that the initialization of the weights plays an important role in the weights that we eventually converge to~\cite{pmlr-v9-sutskever10a}. Therefore, we initialize $w_i^{(0)}$ with the explanation scores from GNNExplainer, i.e., $w_i^{(0)}$ is the average score for relation $R_i$ over all ${\bf S}\in\mathcal{S}$.

\subsection{Uncertainty Estimation}

We use Belief Propagation (BP)~\cite{yedidia&al01} to estimate the uncertainty in an explanation from the factor graph. Specifically, BP is a message-passing algorithm that uses sum-product computations to estimate probabilities. Specifically, the idea is that a node computes the product of all messages coming into it, and sums out itself before sending its message to its neighbors. In a factor graph, there are two types of messages, i.e., messages from variable nodes to factor nodes and messages from factor nodes to variable nodes. The messages from variable to factor nodes involves only a product operation and the messages from the factor to variable nodes involve both a sum and product operation.
\begin{equation}
    v_{var(i)\rightarrow fac(s)}(x_i) \propto \prod_{t\in N(i)\setminus s}\mu_{fac(t)\rightarrow var(i)}(x_i)
\end{equation}
\begin{align}
    \mu_{fac(s)\rightarrow var(i)}(x_i) \propto \nonumber\\ & \sum_{x_{N(s)\setminus i}}f_s(x_{N(s)})\nonumber\\&\prod_{j\in N(s)\setminus i}v_{var(j)\rightarrow fac(s)}(x_j)
\end{align}
where $var(i)$ represents a variable node $i$, $N(i)\setminus s$ represents all the neighbors of $i$ except $s$, $f_s$ is a factor node. Thus, each variable node multiplies incoming messages from factors and passes this to other factor nodes. The factor nodes multiply the messages with its factor function and sums out all the variables except the one that the message is destined for. In practice, the messages are normalized to prevent numerical errors. The message passing continues until the messages converge. In this case, we say that the factor graph is calibrated and we can now derive marginal probabilities from the calibrated factor graph by multiplying the converged messages coming into a variable node. Specifically,
\begin{equation}
    p(x_i)\propto \prod_{t\in N(i)}\mu_{fac(t)\rightarrow var(i)}(x_i)
\end{equation}
We estimate uncertainty of an explanation from the calibrated factor graph. Specifically, let $Q_1\ldots Q_k$ denote relations in the explanation $E(\mathcal{G},\Phi,Y)$. We estimate probabilities from the factor graph denoted by $\mathcal{F}$ that we learn from the symmetric CREs $\mathcal{S}$. Intuitively, a relation $Q_i$ is important in $E(\mathcal{G},\Phi,Y)$ if it is important in $\mathcal{S}$. To quantify this, we compute the change in distributions when we re-calibrate $\mathcal{F}$ by adding new factors obtained from the explanation $Q_1\ldots Q_k$.

\begin{figure*}
\centering
    \subfigure[]{\includegraphics[scale=1.2]{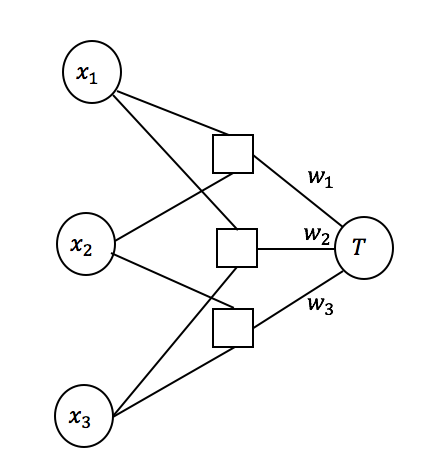}}\quad
    \subfigure[]{\includegraphics[scale=1.2]{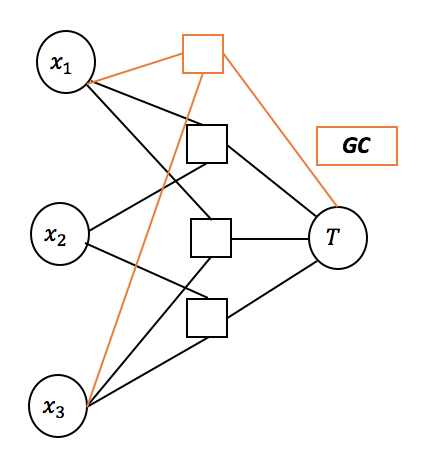}}\quad
    \subfigure[]{\includegraphics[scale=0.6]{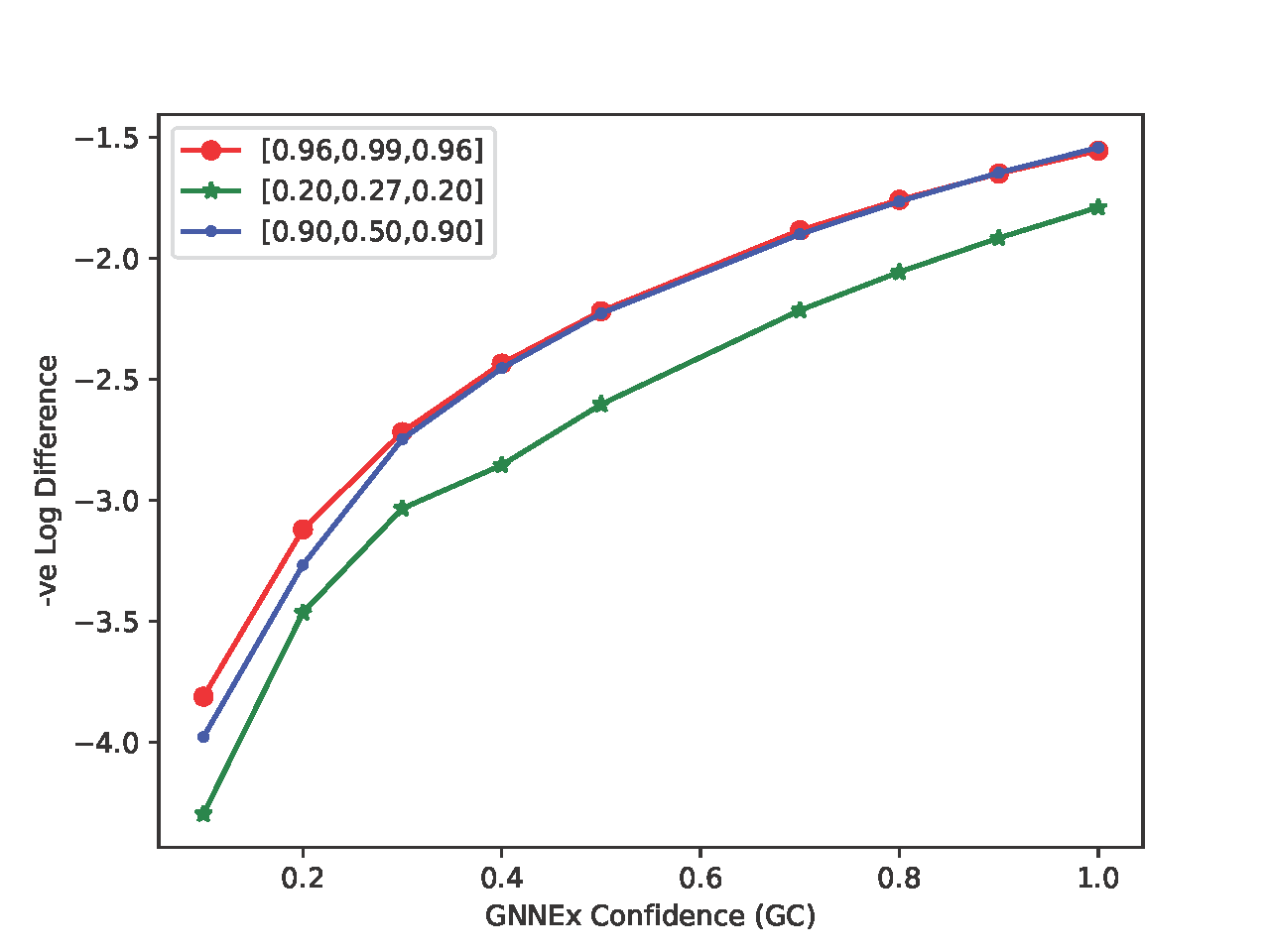}}
\caption{ (a) shows the original factor graph, (b) shows the added factor based on a new explanation $x_1,x_3$ with confidence $GC$ (c) shows the difference in joint probabilities $p(x_1,x_3)$ before and after the factor is added for different values of $GC$.}
\label{fig:fgex}
\end{figure*}

We illustrate our approach with an example in Fig.~\ref{fig:fgex}. As shown here, we have a factor graph with three factors, where each factor explains the influence of a pair of nodes on the target. The weights of the factors ($w_1,w_2,w_3$) encode the uncertainty in the relationship specified by the factors. To quantify the uncertainty of a relation between say $x_1,x_3$ in explaining $T$, we obtain the joint distribution over $P(x_1,x_3)$ after calibration using belief propagation. 
Now, suppose the GNNExplainer gives us an explanation for $T$ that specifies a single relation between $x_1,x_3$ with a confidence equal to $GC$. Our goal is to quantify the reduction in uncertainty given this new explanation. To do this, we add a new factor with weight $GC$ and re-calibrate to obtain a modified distribution $\hat{P}(x_1,x_3)$. 

The graph in Fig,~\ref{fig:fgex} (c) shows the -ve log difference between the two distributions (we show it for the case $x_1=x_3=1$). A larger value indicates that the reduction in uncertainty is larger. As seen here, if the prior uncertainty is high (the green plot), then, it requires $GC$ to be very large (over 90$\%$) to achieve the same level of reduction in uncertainty as is the case for a much smaller $GC$ (around 60$\%$) when the prior uncertainty is low (the red/blue plots).
Further, even though the weight vector for the red plot has larger values than those in the blue plot, we see that the reduction in uncertainty is comparable over all values of $GC$. This is because for the red plot, all the weights are large and therefore, the relationship between $x_1,x_3$ is not significantly more important than the other relationships. Thus, additional knowledge that $x_1,x_3$ is an explanation does not result in a large reduction in uncertainty. On the other hand, as seen by the weights in the blue plot, the factor encoding the relationship between $x_1,x_3$ has a much larger weight compared to that for $x_2,x_3$. Thus, there is a significant difference when the target is explained with a relation between $x_1,x_3$ as compared to something else, say $x_2,x_3$. Therefore, knowledge that a relation between $x_1,x_3$ is the explanation will greatly reduce uncertainty. 

Formally, let us assume that at most $k$ new factors $\hat{t}_1\ldots\hat{t}_k$ are introduced by the GNNExplainer's explanation whose scope has the variable $x_i$. Each of these factors has a weight equal to the confidence (a value between 0 and 1) assigned by GNNExplainer which quantifies its confidence that the corresponding relation is an explanation for the target variable. The new messages are as follows.
\begin{align}
    v_{var(i)\rightarrow fac(s)}(x_i) \propto \nonumber\\
    &\prod_{j=1}^k \mu_{fac(\hat{t}_j)\rightarrow var(i)}(x_i)\nonumber\\
    &\prod_{t\in N(i)\setminus s}\mu_{fac(t)\rightarrow var(i)}(x_i)
\end{align}
The marginal after re-calibration is given by,
\begin{align}
    p(x_i)\propto \nonumber\\
    &\prod_{j=1}^k \mu_{fac(\hat{t}_j)\rightarrow var(i)}(x_i)\nonumber\\
    &\prod_{t\in N(i)}\mu_{fac(t)\rightarrow var(i)}(x_i)
\end{align}
Thus, as the confidence values grow larger, then the messages get amplified in each iteration of BP and the uncertainty reduces since the marginal probability becomes larger. In our case, we store the joint probabilities for all related entities in the symmetric CRE set. Let $p(x_i,x_j)$ be the joint distribution computed for related entities $x_i,x_j$ after calibration. Let $\hat{p}(x_i,x_j)$ denote the joint distribution after re-calibration upon adding factors based on the GNNExplanation. We compute the difference between $p(x_i,x_j)$ and $\hat{p}(x_i,x_j)$ based on the logical structure of the factors. Specifically, since we assume that the structure is a conjunction over $x_i\wedge x_j\wedge Y$, where $Y$ is the explanation target. We compute the average difference over all cases where the formula is satisfied. In the binary case, this corresponds to $x_i=x_j=T=1$. This difference is a measure of reduction in uncertainty when $x_i,x_j$ is related in the explanation.

\begin{algorithm}[]{
%\label{alg:cnnlearn}
\small
\linesnumbered
\caption{Uncertainty Quantification}
\KwIn{GNNExplanation $E(\mathcal{G},\Phi,Y)$ with confidences ${\bf GC}$}
\KwOut{$\delta_{x_i,x_j}$, the uncertainty for the relation connecting $x_i,x_j$ in $E(\mathcal{G},\Phi,Y)$}
\tcp{Computing the symmetric CREs}
Initialize rank $r$\\
Initialize $err$ as $\infty$\\
\While{$err$ $<$ $t$}{
\tcp{Low rank approximation for $\mathcal{G}$ with rank $r$}
    {\bf P} $=$ Adjacency matrix for $\mathcal{G}$\\
    Factorize  {\bf P} into ${\bf Q},{\bf R}$ with $r$ Boolean patterns\\
    $\hat{\bf P}$ $=$ ${\bf Q}^\top{\bf R}$\\
    $\hat{\mathcal{G}}$ $=$ Relational graph with adjacency matrix $\hat{\bf P}$\\
   $\mathcal{S}$ $=$ $\mathcal{S}$ $\cup$ $E(\hat{\mathcal{G}},\Phi,Y)$\\
    Increment $r$\\
    $err$ $=$ $|{\bf P} - \hat{\bf P}|$\\
}
\tcp{Factor Graph}
$FG$ $=$ Factor Graph of $\mathcal{S}$\\
Initialize weights of $FG$ using GNN confidences in explanations for $\mathcal{S}$\\
Learn weights of $FG$ using gradient descent\\

\tcp{Calibration}
$BP(FG)$ $=$ Calibrated FG\\
$p(x_i,x_j)$ $=$ Joint distribution using $BP(FG)$\\
\For{each $(x_i,x_j)$ $\in$ $E(\mathcal{G},\Phi,Y)$}{
Add factor over $(x_i,x_j,T)$ to $FG$\\
}
$\hat{BP}(FG)$ $=$ Re-calibrate FG\\
$\hat{p}(x_i,x_j)$ $=$ Joint distribution using $\hat{BP}(FG)$\\
return $p(x_i,x_j)-\hat{p}(x_i,x_j)$
}
\end{algorithm}

Algorithm 1 summarizes our full approach. Our input is the GNNExplainer's explanation for a relational graph $\mathcal{G}$, using DNN $\Phi$ for target $Y$. Our output is a measure for the reduction in uncertainty when relation $x_i,x_j$ $\in$ $E(\mathcal{G},\Phi,Y)$ is part of an explanation for $Y$. We start by computing the low rank approximations from the input relational graph $\mathcal{G}$. We then explain each of the CREs with GNNExplainer to obtain the set of symmetric CREs $\mathcal{S}$. Next, we construct a factor graph $FG$ from $\mathcal{S}$ and learn its weights. We calibrate $FG$ using belief propagation and obtain the joint distribution $p(x_i,x_j)$. We then add factors to $FG$ from relations in $E(\mathcal{G},\Phi,Y)$ with weights equal to the explanation confidence assigned to the relations. We then re-calibrate the changed $FG$ and compute $\hat{p}(x_i,x_j)$. We return the difference between $p(x_i,x_j)$ and $\hat{p}(x_i,x_j)$.

\section{Experiments}

\subsection{Evaluation Procedure}

We evaluate our approach by measuring the significance of the uncertainty measures that we obtain for an explanation. Specifically, we learn a GCN for node prediction from the input graph and explain target nodes using GNNExplainer on the learned GCN. For each explanation, we apply our approach to compute the uncertainty scores for the relations in the explanation. We then modify the original graph based on these scores and observe changes in predictions made by the GCN. Specifically, we use an approximate statistical test known as the McNemar's test~\cite{McNemar47samplingIndependence} for quantifying differences in prediction. We do this to avoid Type I errors, i.e., errors made by an approximate statistical test where a difference is detected even though no difference exists. In the well-known work by Dietterich~\cite{dietterich1998approximate}, it is shown that McNemar's test has a low Type I error. We run this test as follows. Let $\mathcal{G}$ be the original relational graph and $\Phi$ be the GCN learned from $\mathcal{G}$. We run our approach to obtain uncertainty measures for explanations in predictions made by $\Phi$ on $\mathcal{G}$ and we remove the relation with the $i$-th highest score (higher score means lower uncertainty that the relation is part of the explanation) from $\mathcal{G}$ for each of the target nodes. Thus, we get a reduced graph denoted by $\mathcal{G}^{(i)}$. We then learn a new GCN $\Phi^{(i)}$ on $\mathcal{G}^{(i)}$ and compare the predicted values in $\Phi$ with the predicted values in $\Phi^{(i)}$ through the McNemar's test. If the removed relations are significant, then we would observe a higher score in the McNemar's test along with a small p-value that rules out the null hypothesis that there is no significant difference in predictions made by $\Phi$ and $\Phi^{(i)}$.

\begin{figure*}
    \centering
    \subfigure[]{\includegraphics[scale=0.53]{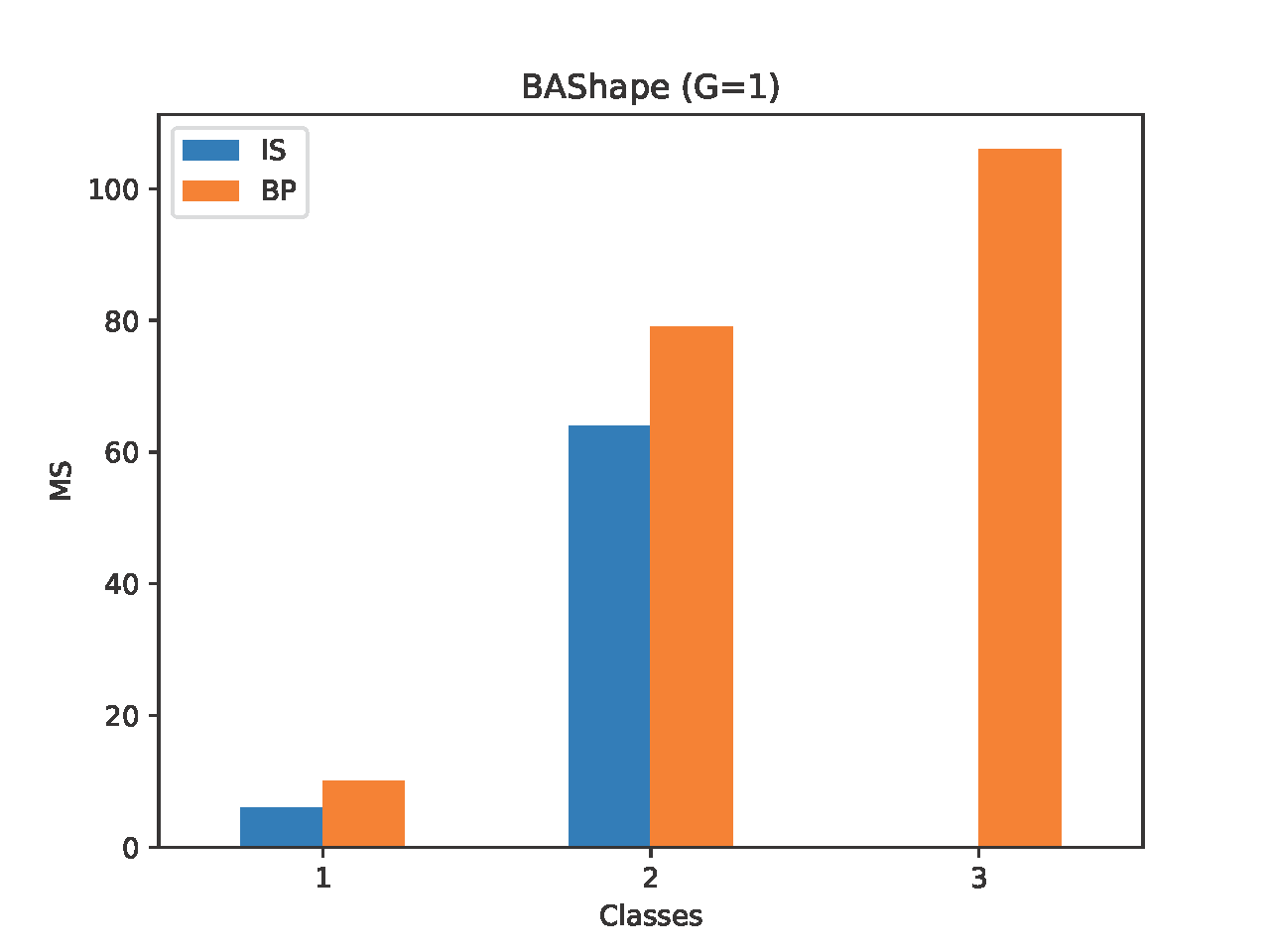}}\quad
    \subfigure[]{\includegraphics[scale=0.53]{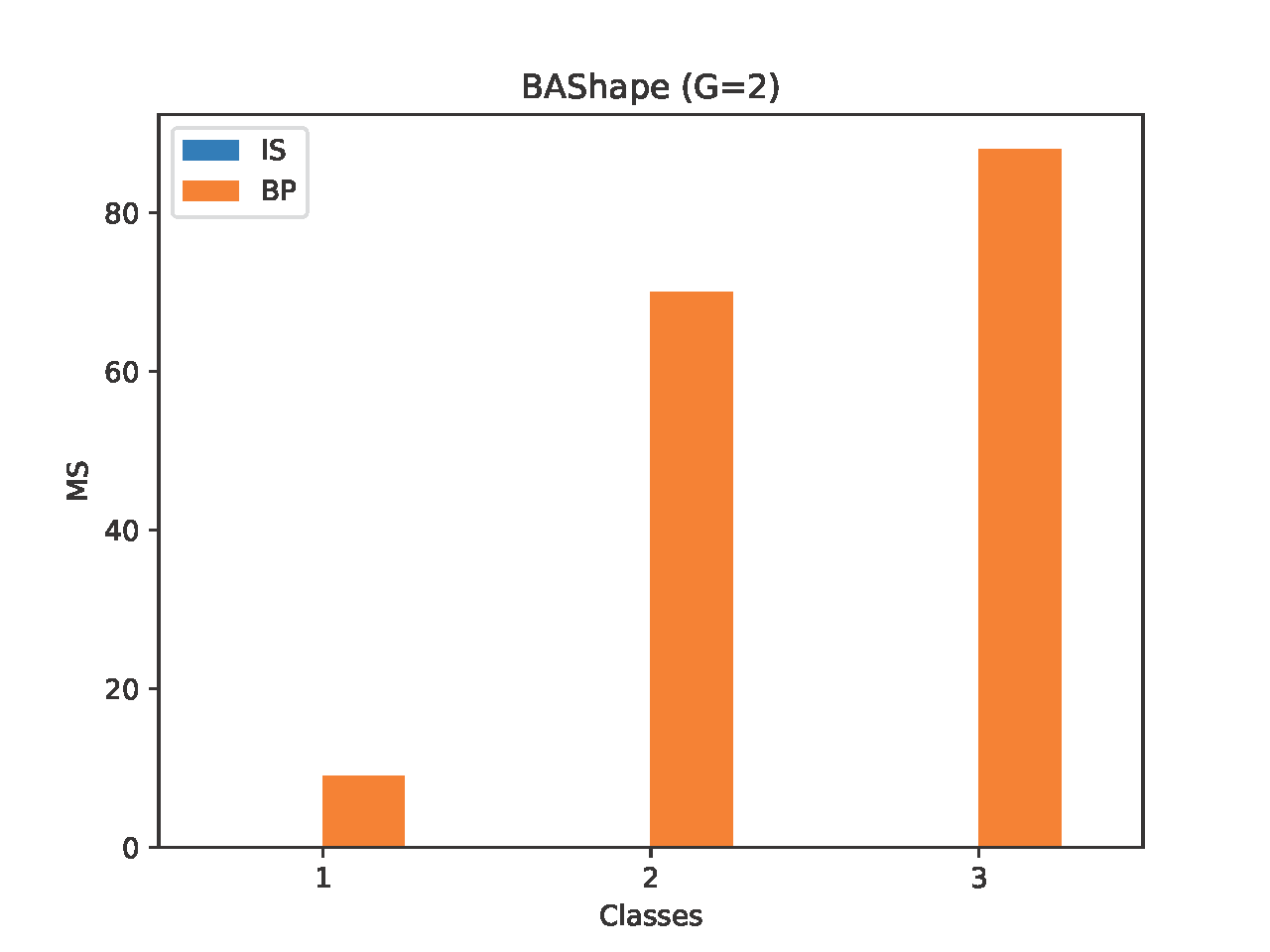}}\quad
    \subfigure[]{\includegraphics[scale=0.53]{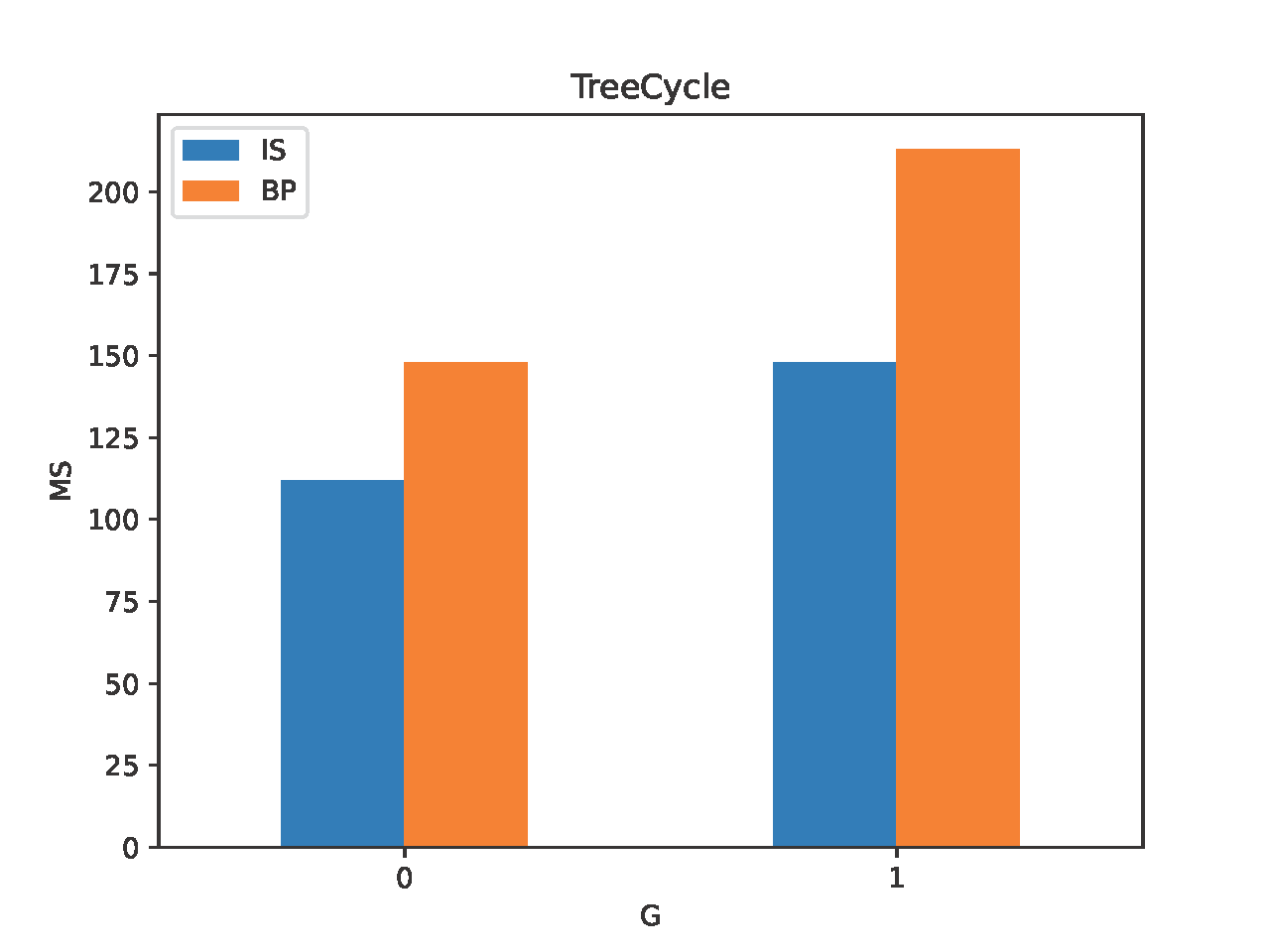}}\quad
    \subfigure[]{\includegraphics[scale=0.53]{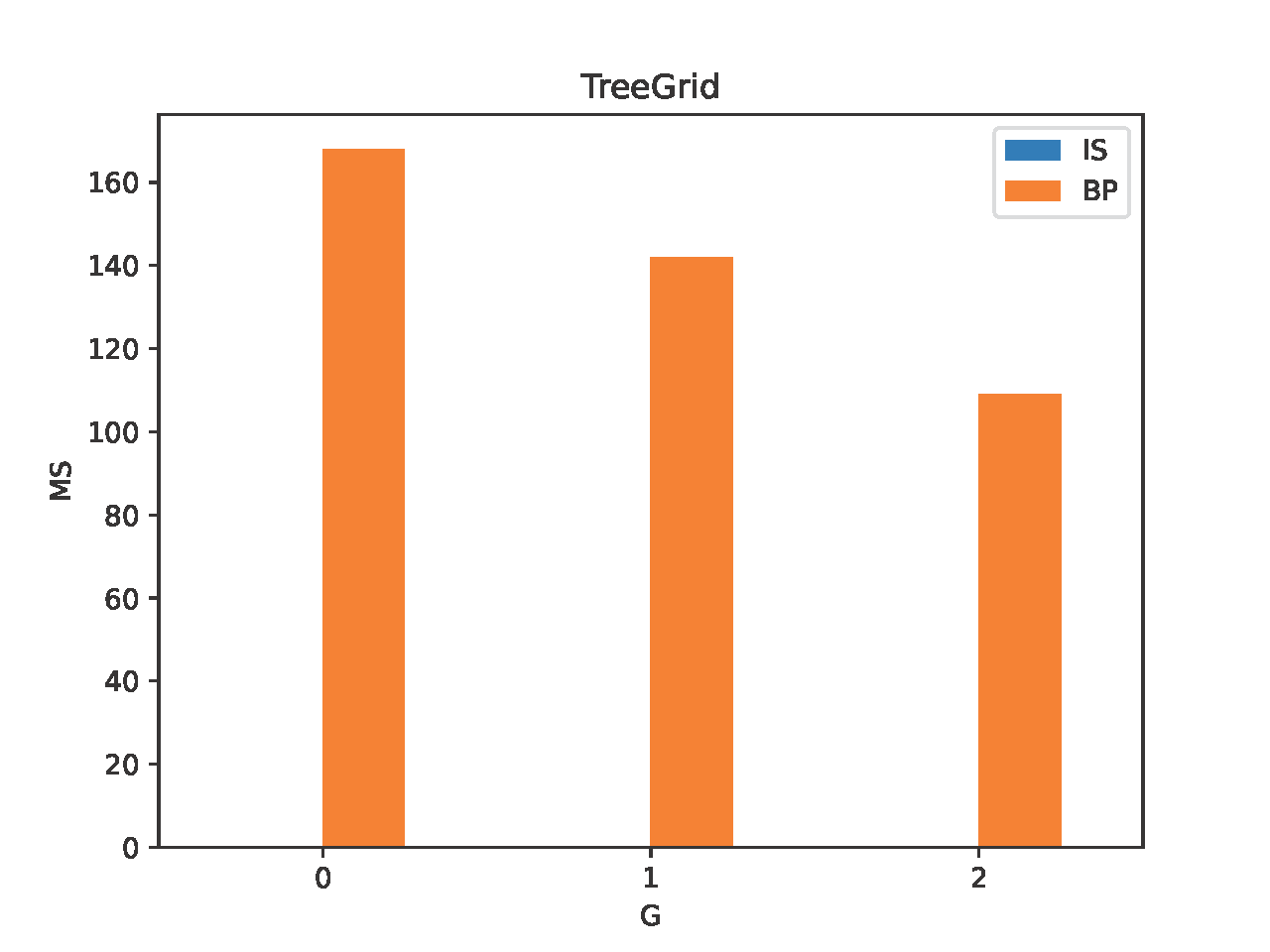}}\quad
    \subfigure[]{\includegraphics[scale=0.53]{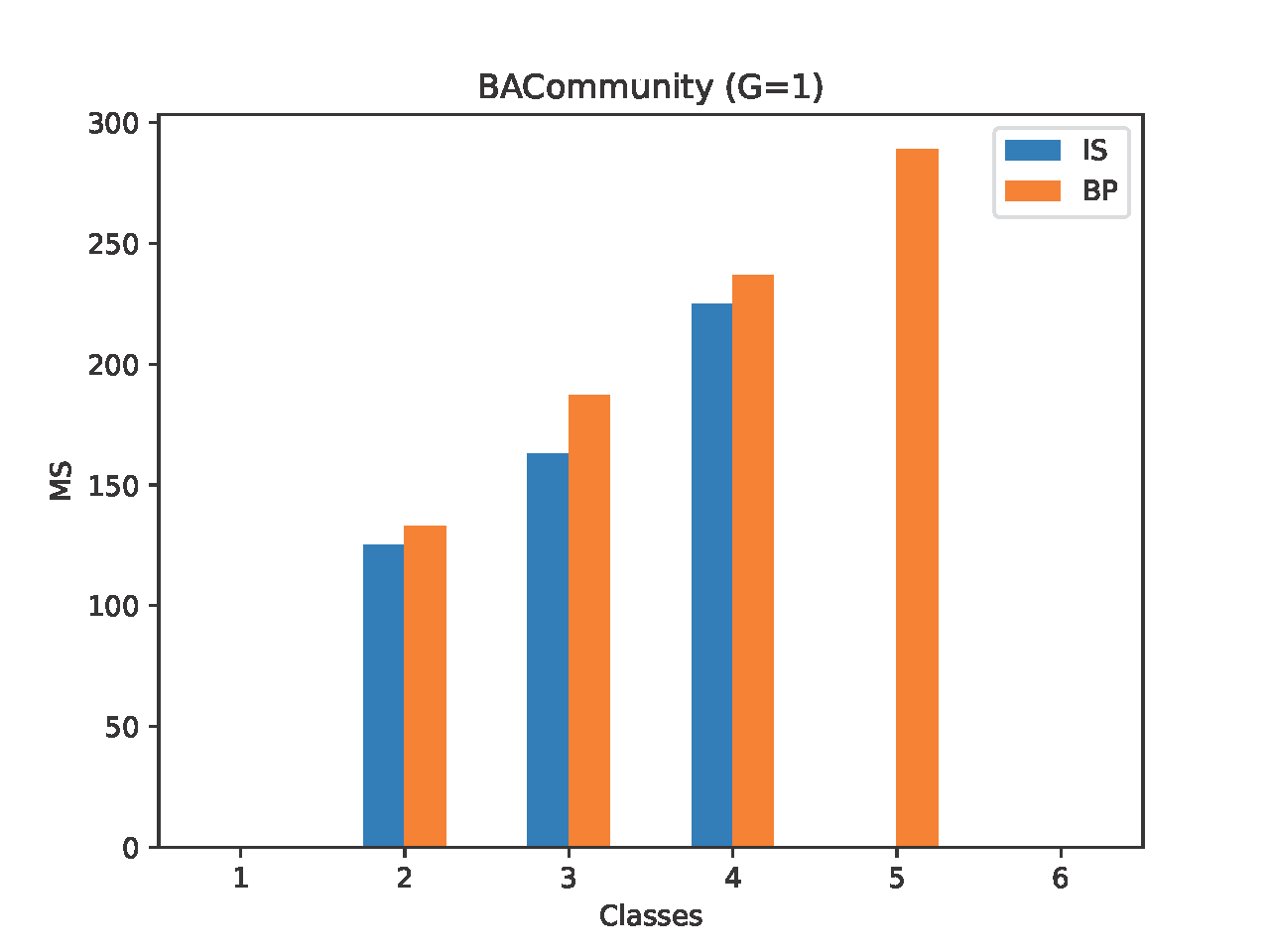}}\quad
    \subfigure[]{\includegraphics[scale=0.53]{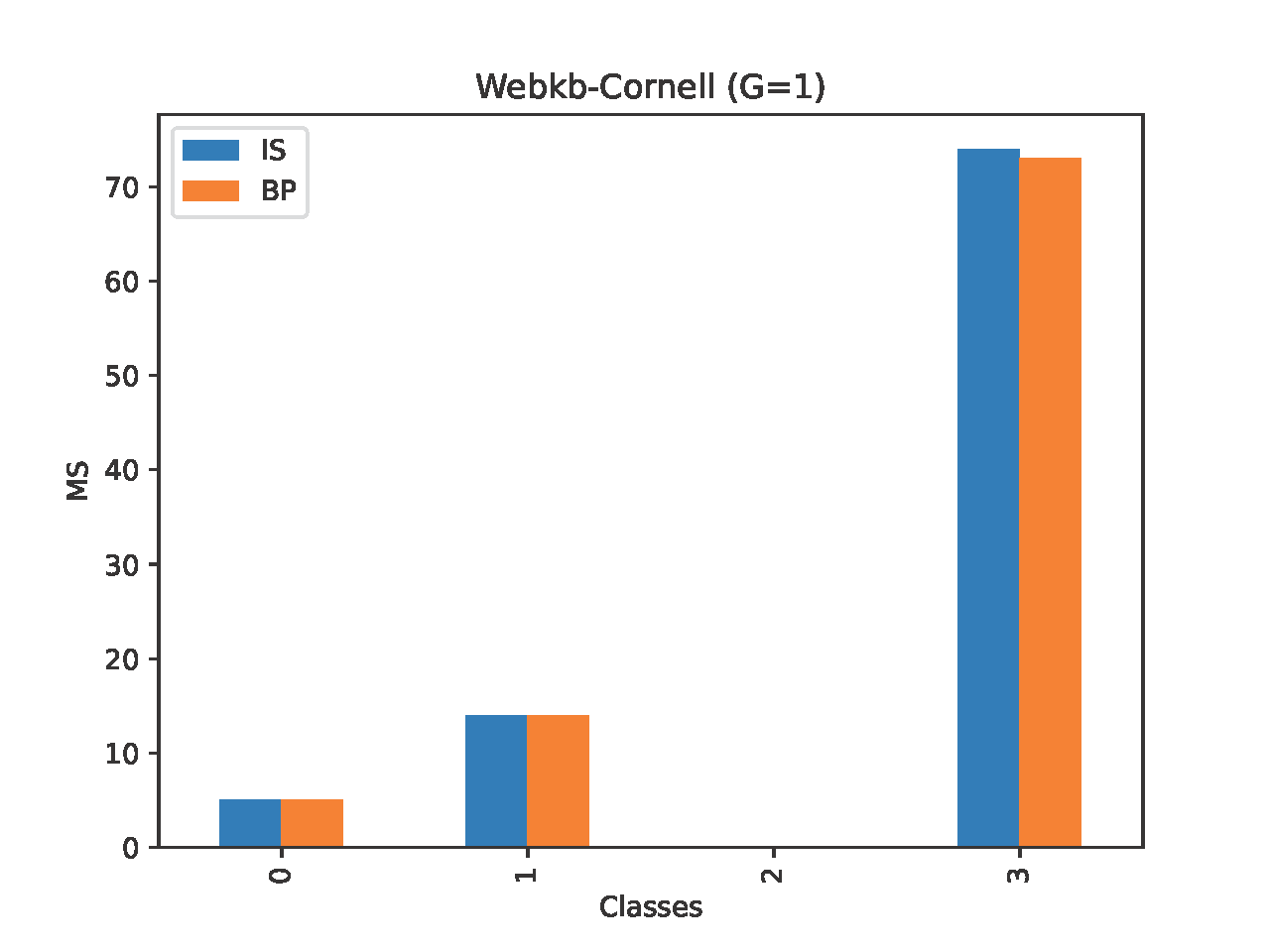}}\quad
    \subfigure[]{\includegraphics[scale=0.53]{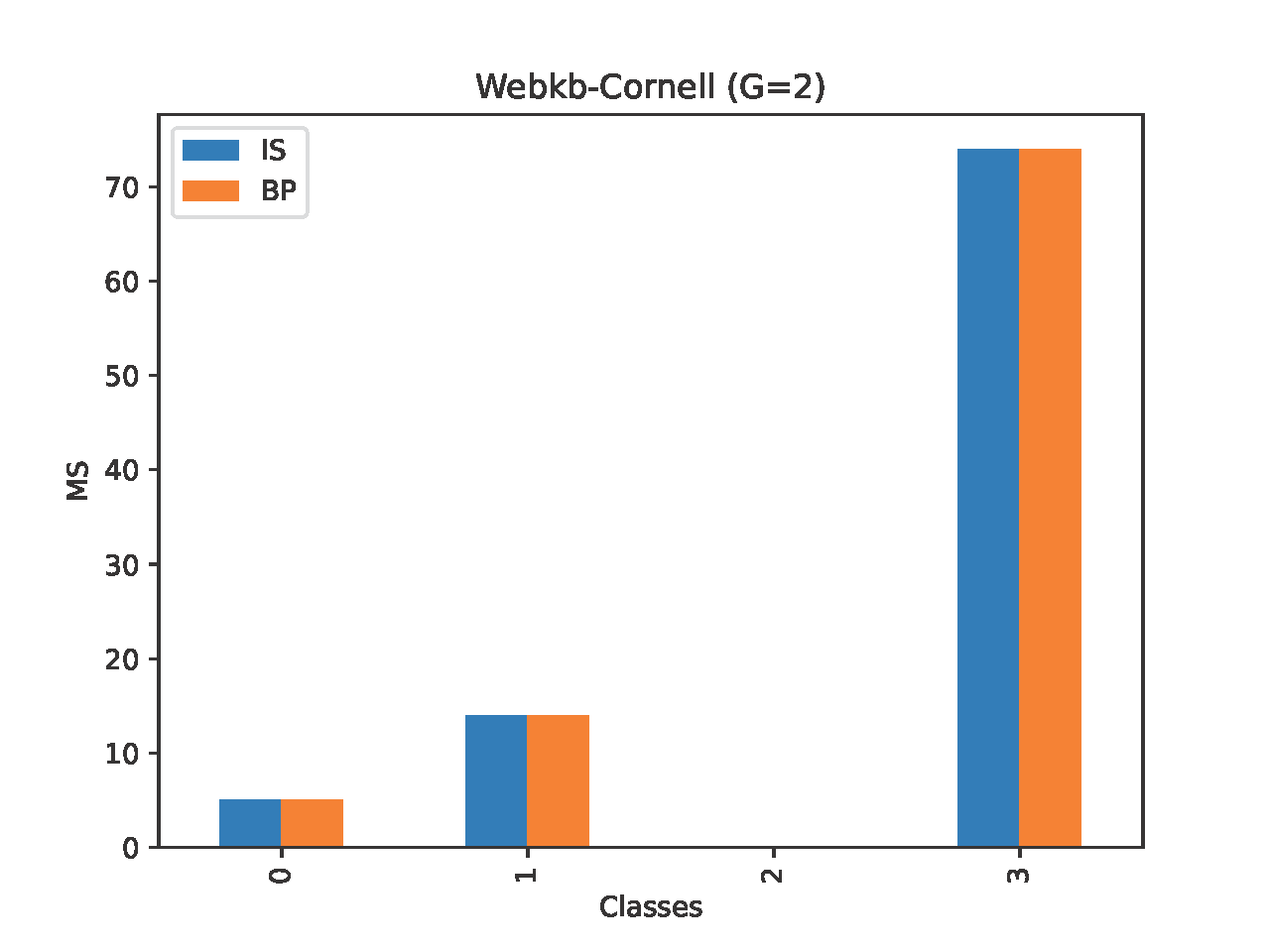}}\quad
    \subfigure[]{\includegraphics[scale=0.53]{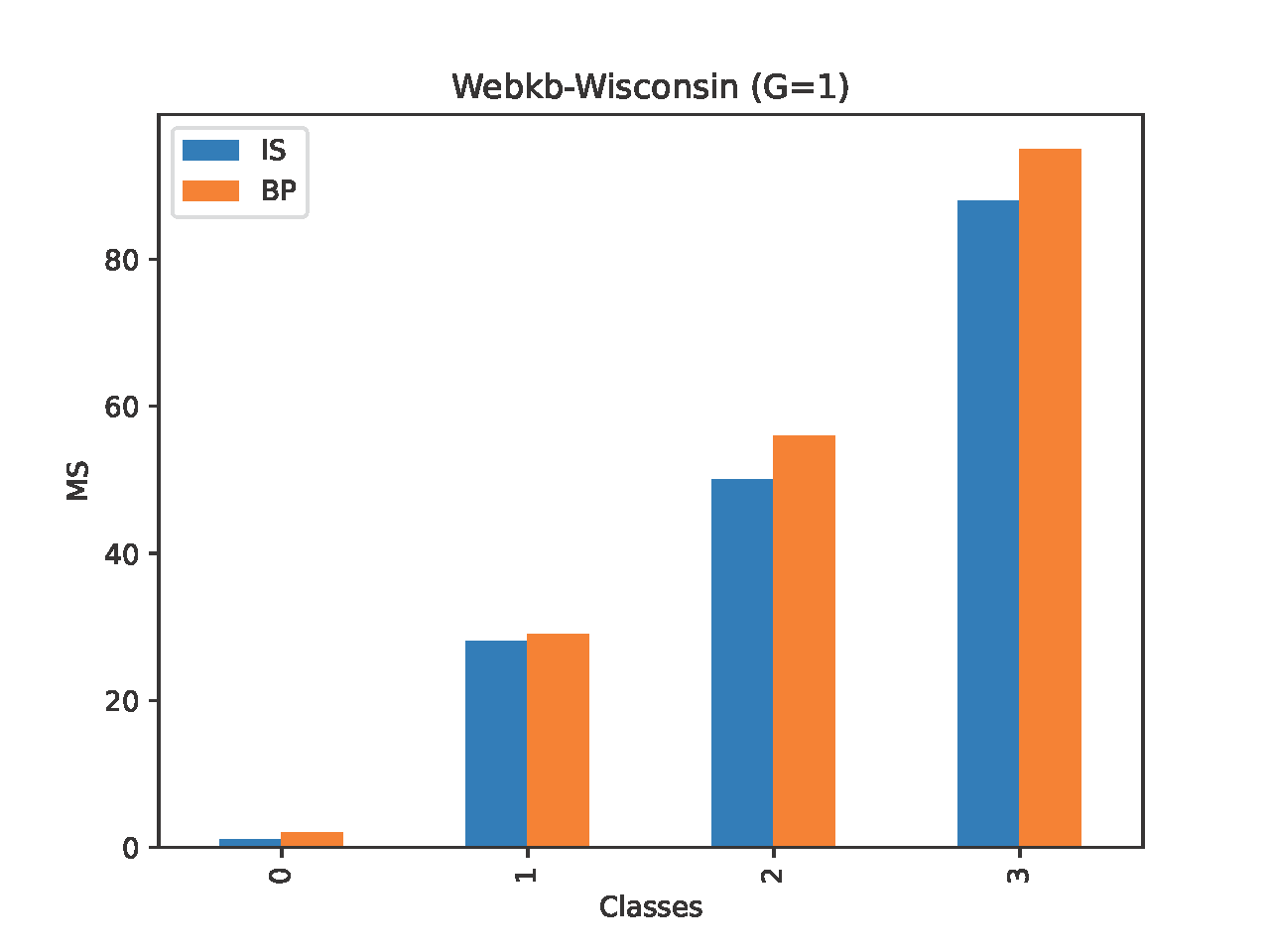}}\quad
    \subfigure[]{\includegraphics[scale=0.53]{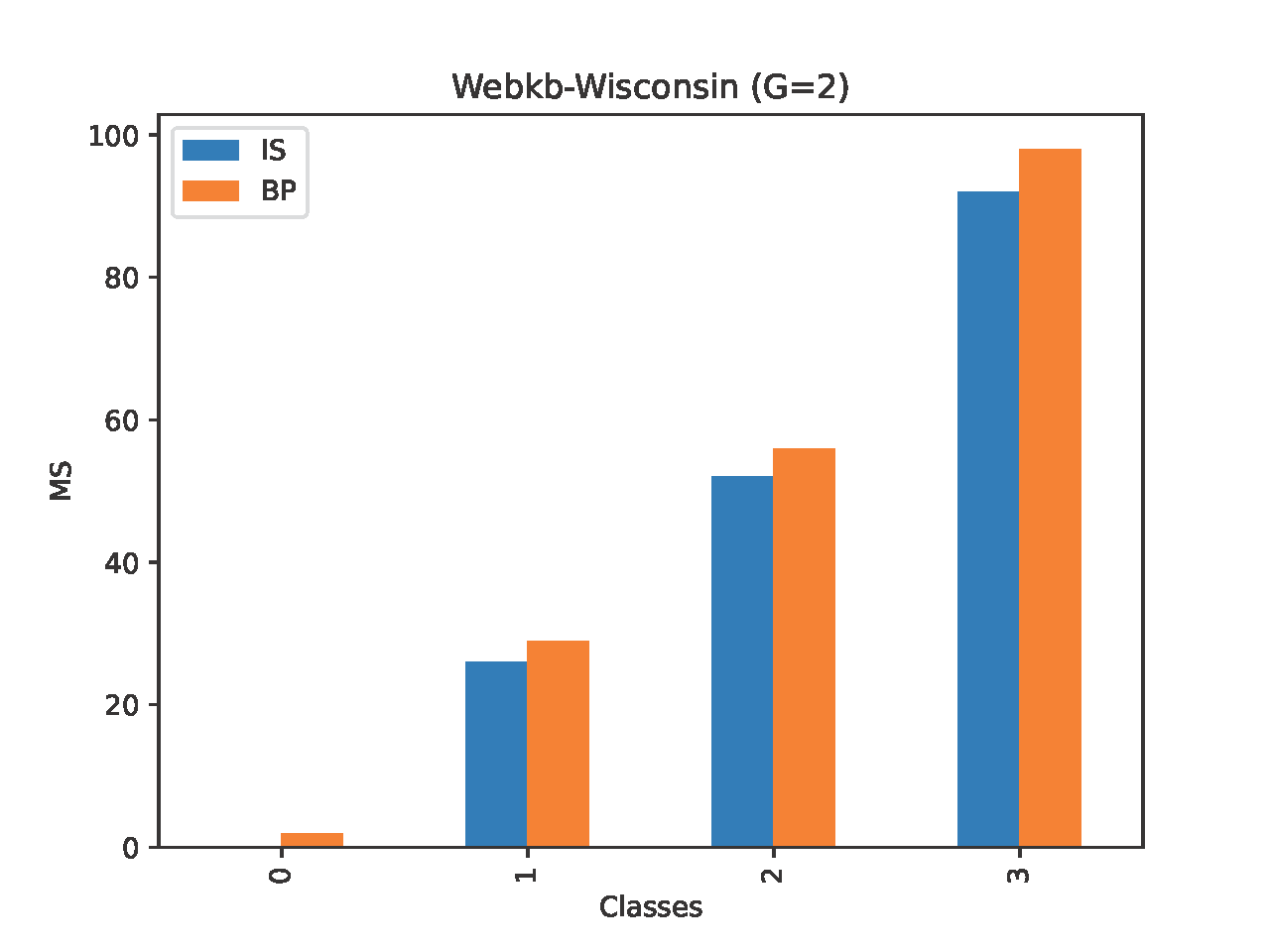}}
   % \subfigure[]{\includegraphics[scale=0.35]{plotsbp/texas-1.pdf}}\quad
    \caption{Results from McNemar's test to verify uncertainty quantification in benchmarks.}
    \label{fig:results1}
\end{figure*}

\begin{figure*}
    \centering
    \subfigure[]{\includegraphics[scale=0.53]{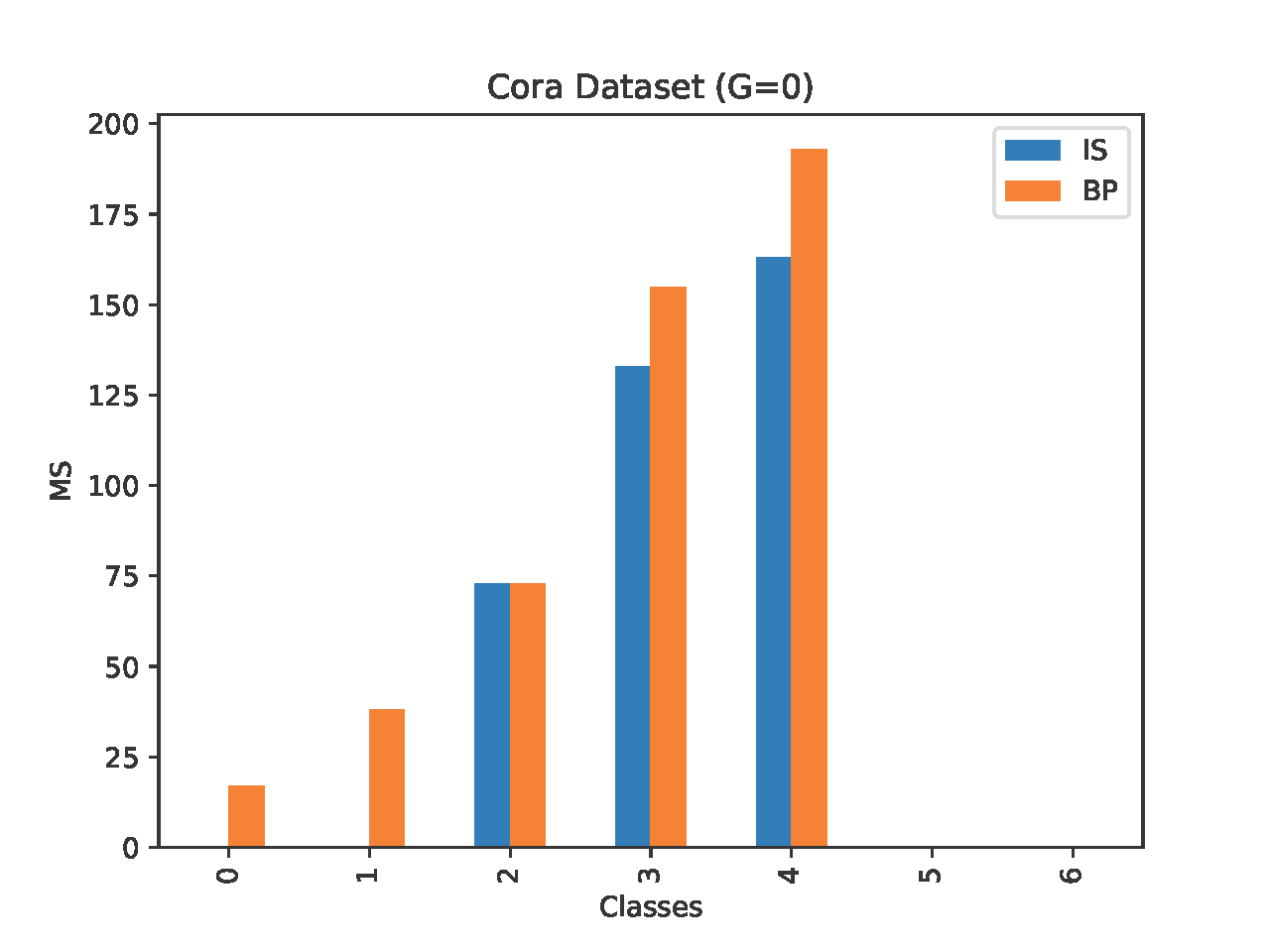}}\quad
    \subfigure[]{\includegraphics[scale=0.53]{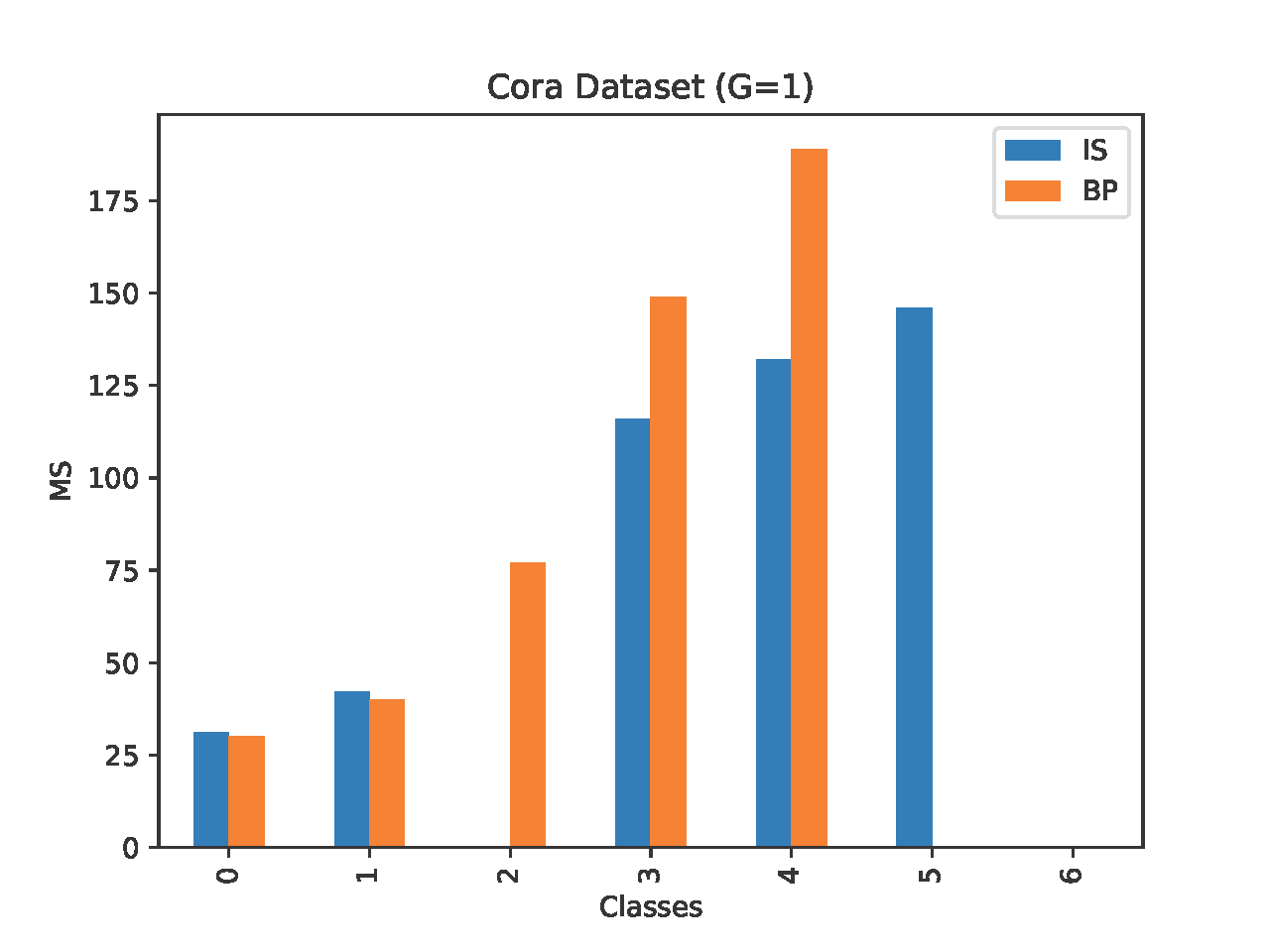}}\quad
    \subfigure[]{\includegraphics[scale=0.53]{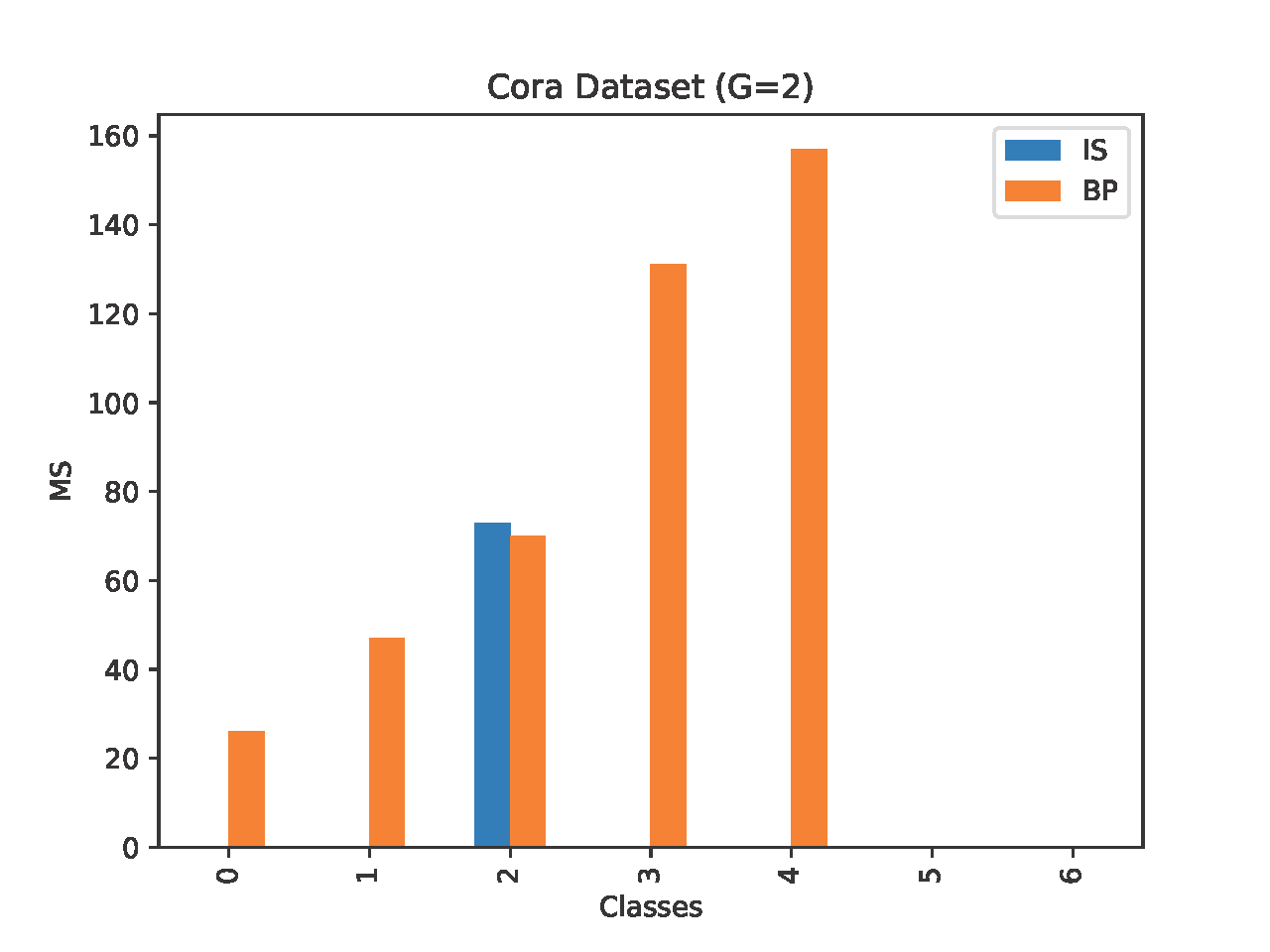}}\quad
    \subfigure[]{\includegraphics[scale=0.53]{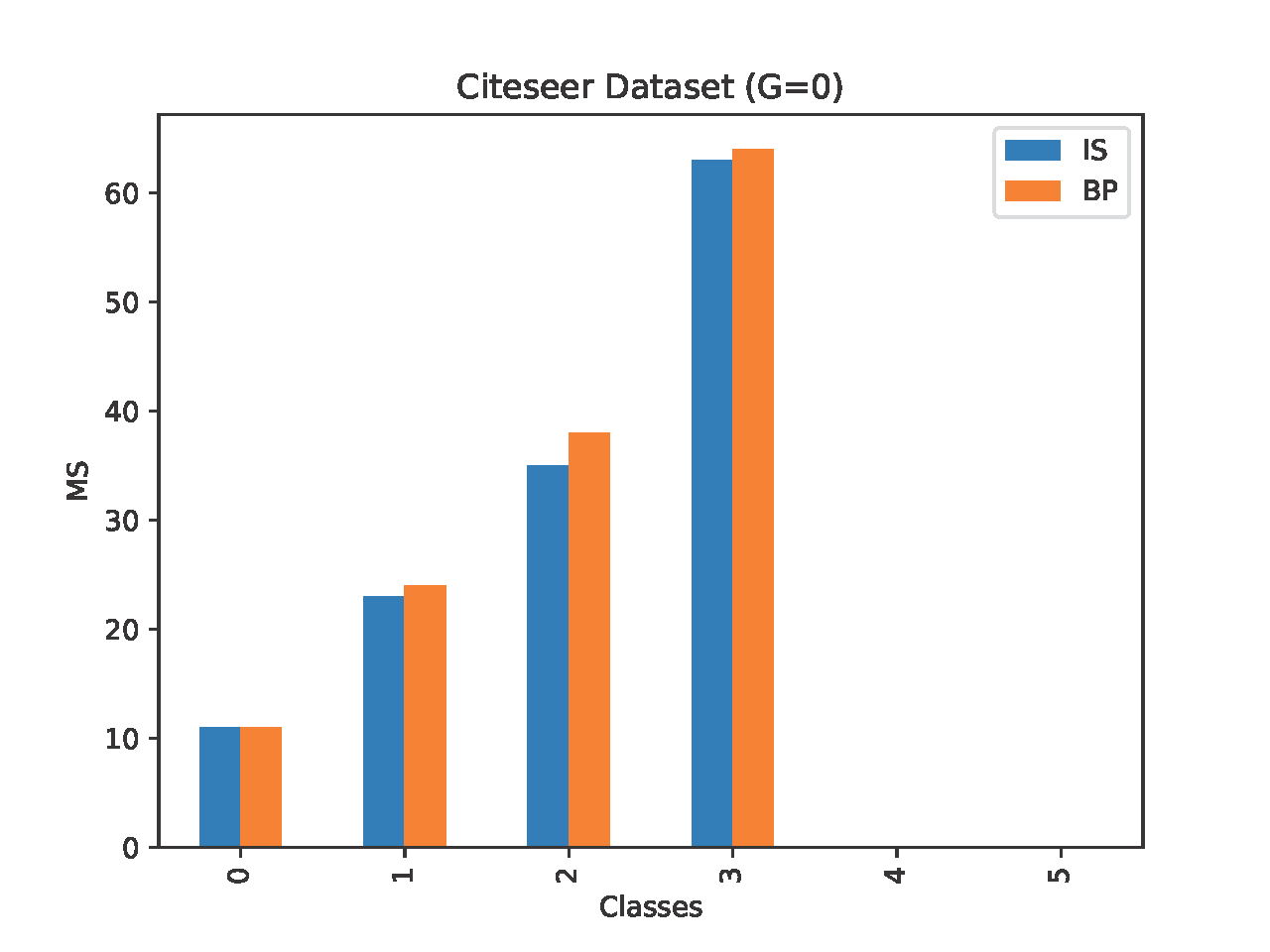}}\quad
    \subfigure[]{\includegraphics[scale=0.53]{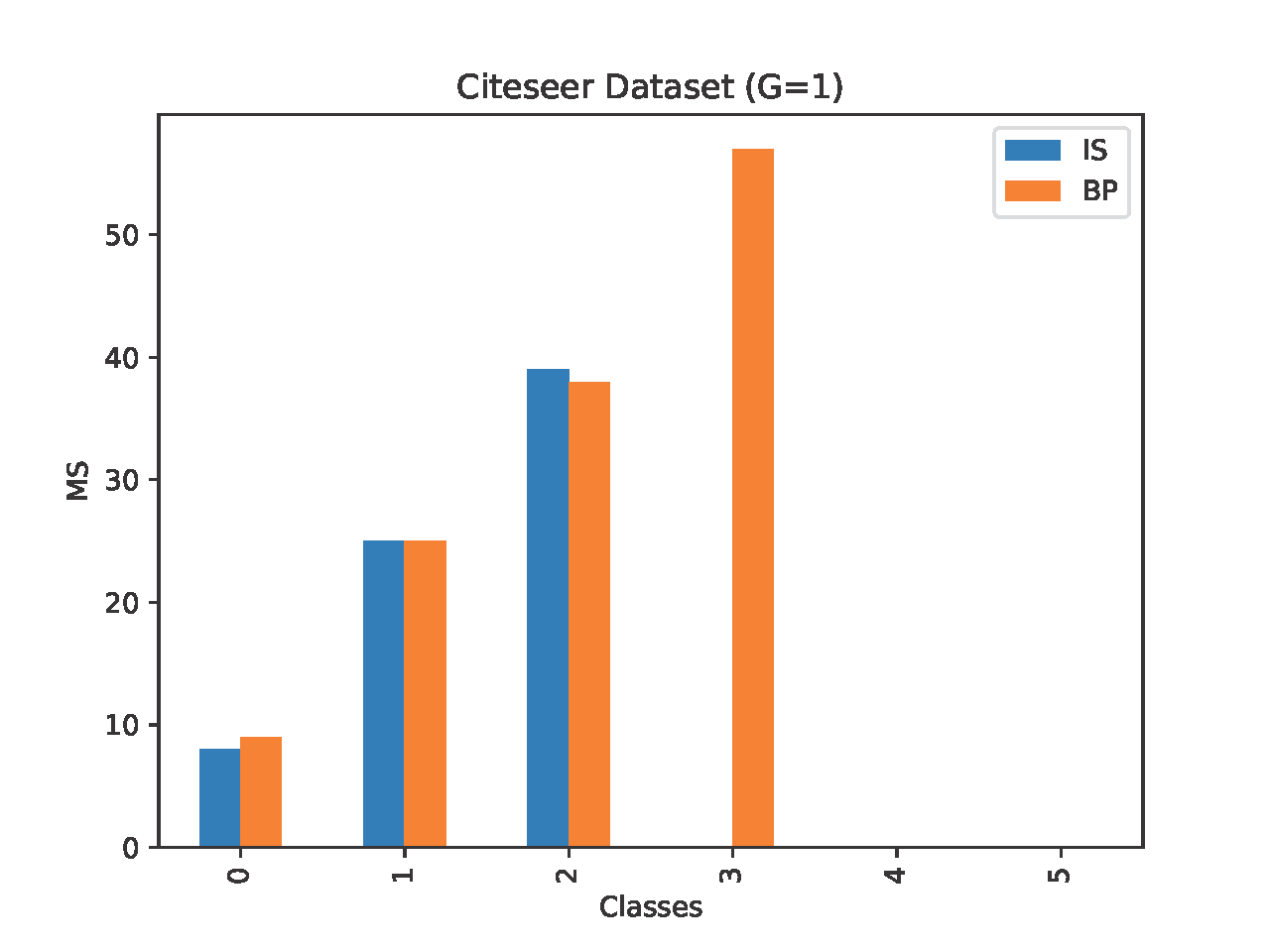}}\quad
    \subfigure[]{\includegraphics[scale=0.53]{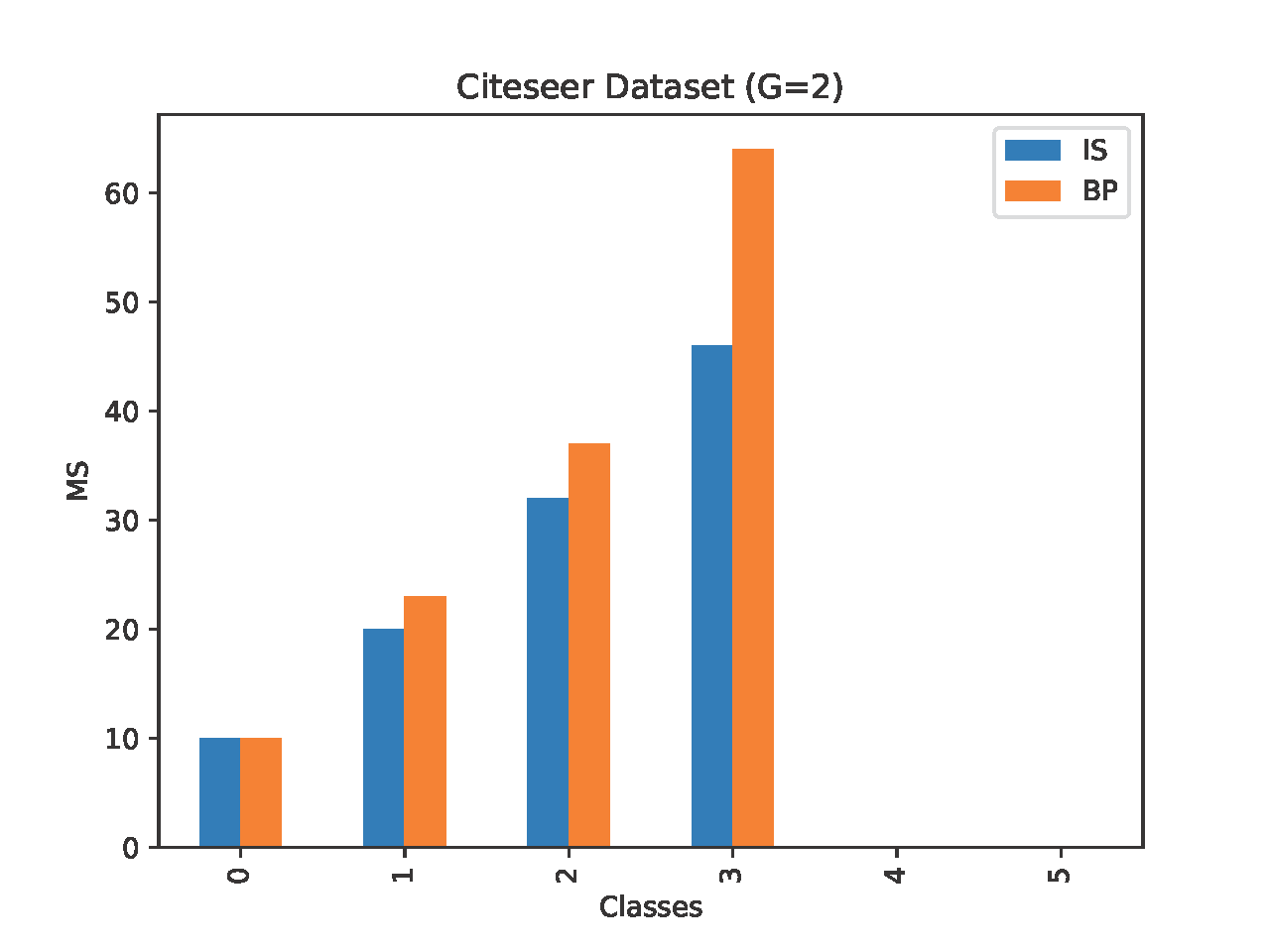}}
    \caption{Results from McNemar's test to verify uncertainty quantification in benchmarks.}
    \label{fig:results2}
\end{figure*}
\subsection{Setup}

We implement our approach using the Deep Geometric Learning (DGL) library in Pytorch. We run all our experiments on a single Tesla GPU machine with 64GB RAM. For the factor graph learning and inference using belief propagation, we used the implementations in the open-source pgmpy library. For the GCN, we varied the hidden dimensions size between 16 and 512, and chose the one with the optimal validation accuracy for a given dataset. We used a maximum of 10K epochs for training the GCN. We used the GNNExplainer from DGL to explain node predictions made by the GCN. We set the number of hops to 2 in the explanations, i.e., any relation that is an explanation for a target node is at most two hops away from the target. For the low rank approximation, we used Nimfa~\cite{JMLR:v13:zitnik12a} which is a python library for Nonnegative Matrix Factorization. We used the default parameters for BMF but set the number of iterations between 10K and 100K depending on the dataset. We initialized the rank to one where the approximation error was less than $25\%$ of the total number of edges in the input graph and stopped increasing the rank if we observed that the rank plateaued or if the approximation error was within $5\%$ of the total number of edges.

% \begin{center}
% \begin{tabular}{||c c c c c c c c c||} 
%  \hline
%  BA-Shape & BA-Community & Tree-cycle & Tree-grid & Cora & Citeseer & Cornell & Texas & Wisconsin \\ [0.5ex] 
%  \hline\hline
%  % \\ [1ex] 
%  % \hline
% \end{tabular}
% \end{center}

\subsection{Datasets}

%We use the benchmark datasets shown in Table~\ref{tab:data} to evaluate our approach. 
We used the following publicly available benchmarks for GNNs: BAShapes, BACommunity, TreeGrids, TreeCycles, Cora, Citeseer and WebKB.
%These benchmarks are widely used in explanations for GNNs and also in relational learning. 
In each case, we compare our results with uncertainty estimates from GNNExplainer. GNNExplainer assigns scores based on the conditional entropy equation Eq.~\eqref{eq:condentropy}. A higher score for a relation indicates greater importance in the explanation. As described in our evaluation procedure, we create a reduced graph based on these scores to compare with our approach. We denote our approach as BP and the GNNExplainer based scores as IS in the results.

If the explanation contains a single node/edge, we filter them from our results. We run the McNemar's test for each class separately and report the statistic values. For the BAShapes, BACommunity, TreeGrid and TreeCycle datasets, we show results for all classes except when the node class is equal to 0 which corresponds to the base nodes. In this case, we explain the nodes from the motifs attached to the base nodes which have non-zero class values. In case the statistic is not significant, i.e., the null hypothesis is true that there is no change in predictions made by $\Phi^{(i)}$ and $\Phi$, we report the statistic as 0. When a class contained less than 10 nodes, we do not report that class in the results since the p-values were not significant. We show the results for $\Phi^{(1)}$, $\Phi^{(2})$, $\ldots$ (labeled as $G=1,G=2,\ldots$ in the graphs) as long as roughly, the same number of edges are removed for both IS and BP. Our data and implementation are available here\footnote{https://github.com/abisha-thapa/explain}.

\subsection{Results}

The results are shown in Fig.~\ref{fig:results1} and Fig.~\ref{fig:results2}. For BAShapes, BP scores higher McNemar’s statistic (MS) values over all classes. Further, using IS, we could not obtain statistical significance for $G=2$ indicating worse uncertainty quantification. For TreeCycle for both $G=1$ and $G=2$, the MS scores for BP were larger than IS. For TreeGrid, we could not obtain statistical significance for any of the values of $G$. This also indicates that as the structure gets more complex (BAShapes is simpler than TreeCycle and TreeGrid), uncertainty quantification becomes more reliable using our approach. For BACommunity, our results were slightly better than IS and also more significant over some classes. For WebKB, we observed that IS and BP had very similar performance. One of the reasons for this is related to the accuracy of the GCN model. The accuracy here was significantly lower (between 50-60$\%$) for WebKB. Thus, it indicates that when the GNN has poor performance, explanations may be harder to verify. For the Cora dataset, BP achieves better MS scores and significance compared to IS for most values of $G$. As $G$ increases, the significance of IS reduces, for instance at $G=3$, most of the values produced by IS were statistically insignificant. Citeseer shows similar results, where for $G=0$, BP and IS gave us similar results, but for larger values of $G$, the MS values for BP were more significant than those for IS.
%Thus to summarize, over most of the tested benchmarks, using BP, we were able to quantify uncertainty in the relations of the explanation in a more statistically significant manner compared to IS.

\section{Conclusion}

%Explanations for relational data are harder to interpret since they involve complex structures (e.g. graphs). 
We developed an approach to verify relational explanations using a probabilistic model. Specifically, we learn a distribution from multiple counterfactual explanations, where a counterfactual representing symmetrical approximation of the graph was learned using low-rank Boolean factorization. From the counterfactual explanations, we learn a factor graph to estimate uncertainty in relations specified by a new explanation.  Our results on several benchmarks show that these estimates are statistically more reliable compared to estimates from GNNExplainer.
In future, we will develop verifications with user-feedback.
%Further, we will also explore specific applications in domains such as education where we can verify context-specific explanations.

\bibliographystyle{IEEEtran}
\bibliography{main}

\end{document}